\documentclass[letterpaper, 10 pt, conference]{ieeeconf}
\IEEEoverridecommandlockouts
\usepackage{amsmath,amsfonts,bm}

\def\1{\bm{1}}

\def\vtheta{{\bm{\theta}}}

\def\vb{{\bm{b}}}

\def\vd{{\bm{d}}}
\def\ve{{\bm{e}}}
\def\vf{{\bm{f}}}

\def\vh{{\bm{h}}}

\def\vp{{\bm{p}}}

\def\vz{{\bm{z}}}

\def\vphi{{\bm{\phi}}}

\def\mE{{\bm{E}}}

\def\mL{{\bm{L}}}

\def\mZ{{\bm{Z}}}

\DeclareMathAlphabet{\mathsfit}{\encodingdefault}{\sfdefault}{m}{sl}
\SetMathAlphabet{\mathsfit}{bold}{\encodingdefault}{\sfdefault}{bx}{n}
\newcommand{\tens}[1]{\bm{\mathsfit{#1}}}

\def\tX{{\tens{X}}}

\def\sC{{\mathbb{C}}}
\def\sD{{\mathbb{D}}}

\def\sY{{\mathbb{Y}}}

\newcommand{\R}{\mathbb{R}}

\usepackage[T1]{fontenc}

\usepackage{graphicx}
\usepackage{stfloats}
\usepackage{float}
\usepackage{graphicx}
\usepackage{amsmath}
\usepackage{bm}
\usepackage{amssymb}
\usepackage{booktabs}

\usepackage{makecell}
\usepackage{subcaption}
\usepackage{multirow}
\usepackage{xcolor}
\usepackage{colortbl}
\usepackage{tcolorbox}
\usepackage{array}
\makeatletter
\let\labelindent\relax
\makeatother
\usepackage{enumitem}
\usepackage{pifont}
\usepackage{tikz}
\usepackage{adjustbox}
\usepackage{marvosym}

\newcommand{\mAP}{\mathrm{mAP}}

\RequirePackage{xspace}
\makeatletter
\DeclareRobustCommand\onedot{\futurelet\@let@token\@onedot}
\def\@onedot{\ifx\@let@token.\else.\null\fi\xspace}
 
\def\ie{\emph{i.e}\onedot} 
\def\cf{\emph{cf}\onedot}

\definecolor{customblue}{HTML}{2B4141}
\definecolor{lightblue}{rgb}{0.9, 0.95, 1.0}
\definecolor{lightpruple}{HTML}{E1D5E7}

\newcommand{\smalltriangleright}{%
  \tikz[baseline=-0.7ex]{
    \fill[customblue, rounded corners=1.2pt] 
      (0,-0.8ex) -- (1.39ex,0) -- (0,0.8ex) -- cycle; 
  }%
  \hspace{0.2em}%
}

\begin{document}

\title{\LARGE \bf MECAIL: Communication-Aware Incremental Learning for\\Object Detection with 14.6 KB Spatiotemporal Experts}

\author{%
    Matthias Neuwirth-Trapp$^{1,2,}$\thinspace\textsuperscript{\Letter},
    Maarten Bieshaar$^{2}$,
    Danda Paudel$^{3}$,\\
    Konrad Schindler$^{1}$,
    Luc Van Gool$^{3}$,
    and Christos Sakaridis$^{1}$%
    \thanks{$^{1}$ETH Zurich, Zurich, Switzerland. \textsuperscript{\Letter}\, mneuwirth@ethz.ch}%
    \thanks{$^{2}$Bosch Research, Hildesheim, Germany.}%
    \thanks{$^{3}$INSAIT, Sofia University ``St.\ Kliment Ohridski'', Sofia, Bulgaria.}%
}

\maketitle

\begin{abstract}
Intelligent transportation systems require Incremental Learning (IL) to continually improve their overall performance in dynamic environments. However, most edge devices lack the computational resources to support on-device IL, requiring updates to be transmitted from centralized servers. We propose using this setup to obtain dense, specialized module coverage that adapts a fixed base model to specific spatiotemporal contexts, such as parking lots, gas stations, ferries, or construction sites. However, in order to reliably transmit these modules to the edge device, using TCP, UDP, and BTP over V2X, Wi-Fi, and 2G-5G hardware, we establish a strict limit of 14.6~KB per module to fit within the first TCP window and to minimize UDP/BTP fragmentation. We further introduce Mixture-of-Experts for Communication-Aware Incremental Learning (MECAIL), the first method that meets this strict requirement, in which each new domain or environment is served by a small expert network that adapts the base model. We validate MECAIL on D-RICO and ODinW-13, where it largely matches the performance of parameter-heavy approaches while enabling practical, bandwidth-efficient large-scale deployment. This allows comprehensive coverage by experts for highly specific, focused, and temporary situations.
\end{abstract}

\begin{figure}[t!]
    \centering
    \begin{subfigure}[b]{1\linewidth}
        \centering
        \includegraphics[width=1\linewidth, clip]{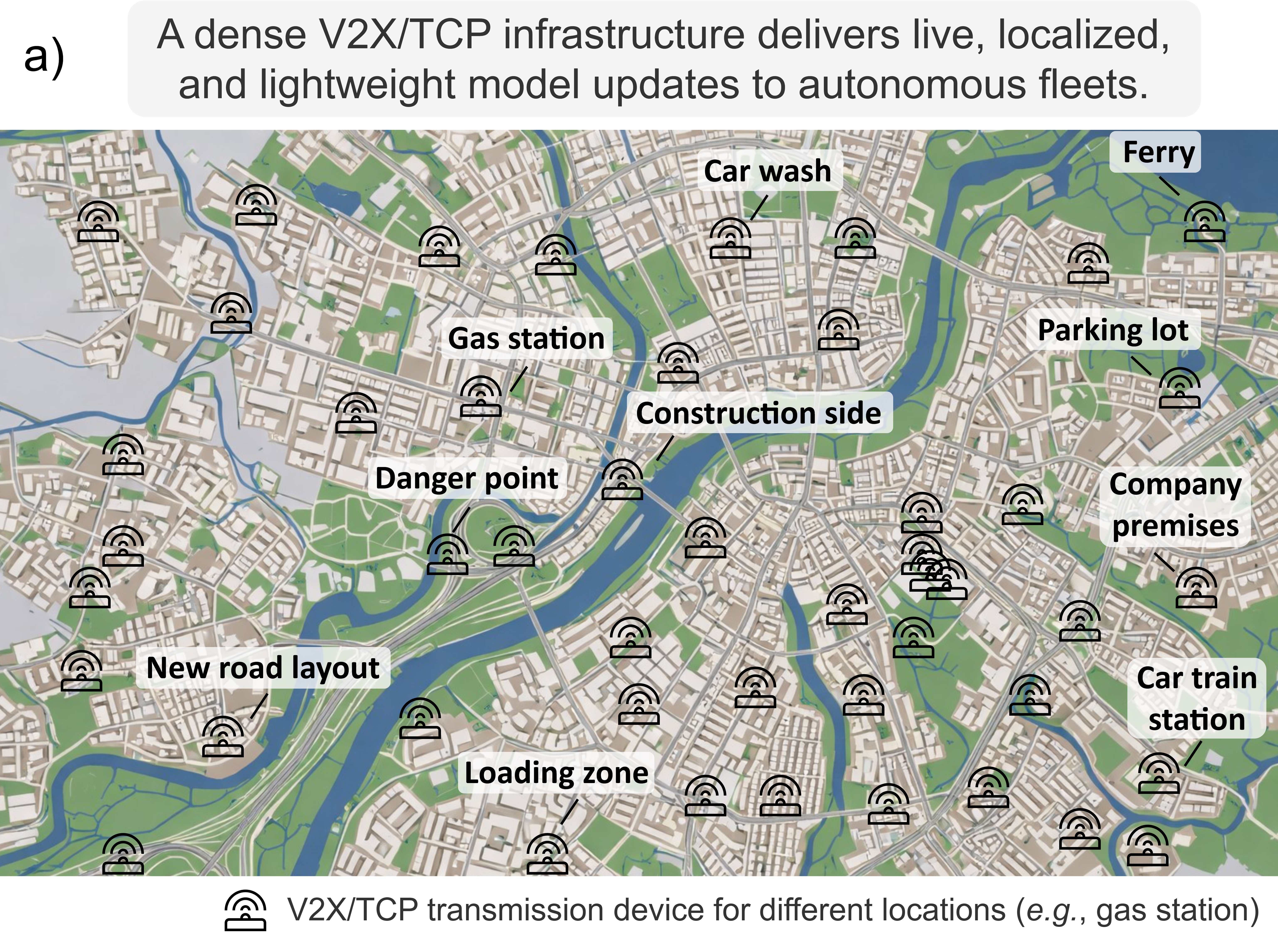}
    \end{subfigure}
    \vspace{0.05cm}
    \begin{subfigure}[b]{1\linewidth}
        \centering
        \includegraphics[width=1\linewidth, clip]{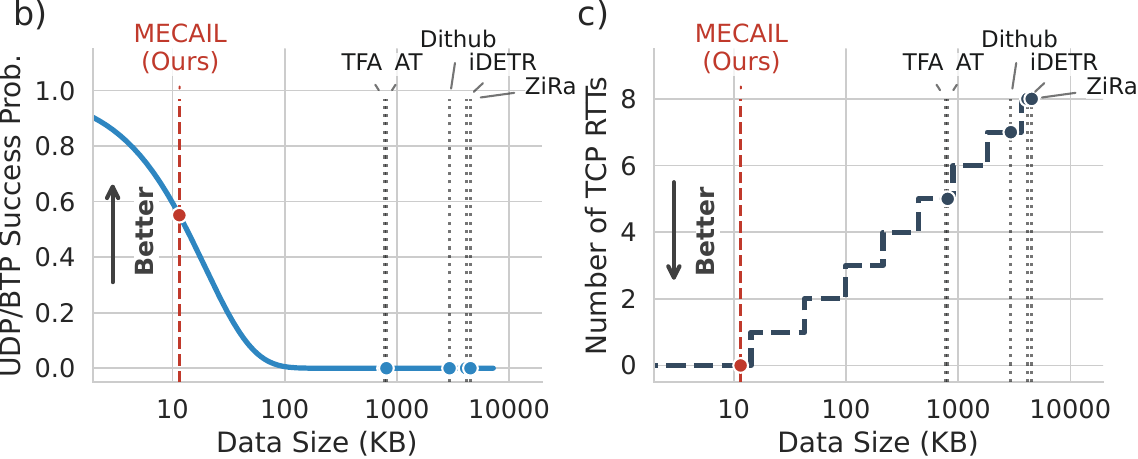}
    \end{subfigure}

\caption{
\textbf{MECAIL spatiotemporal coverage and transmission performance.}
\textbf{(a)} Dense spatiotemporal coverage of experts across diverse contexts, transmitted via V2X, Wi-Fi, or 2G–5G.
\textbf{(b)} V2X transmission success probability for a update (400-byte chunks, 98\% per-chunk success). MECAIL's update size is smaller, reducing UDP/BTP fragmentation and all-or-nothing loss, thereby maintaining high reliability.
\textbf{(c)} Number of TCP round-trips (RTTs) required to deliver the update under Slow Start. Only MECAIL fits within the initial congestion window, enabling 0-RTT delivery, whereas baselines need multiple RTTs.}
    \label{fig1}
    \vspace{-10pt}
\end{figure}

\section{Introduction}
We propose an intelligent transportation system approach that relies on dense expert coverage rather than static, monolithic networks. In this architecture, a generalized onboard base model is dynamically augmented by millions of infrastructure-distributed experts. Consider an autonomous vehicle approaching a complex gas station, a highway construction zone, or an unusual type of ferry (\cf Figure~\ref{fig1}a). Instead of relying on a generalist model, the vehicle receives an \textit{ad hoc} update specifically trained and validated for that spatiotemporal context. This enables instant capability specialization for accident hotspots, temporary environmental changes, fleet-reported uncertainty, or anticipated difficulties.

Formally, this strategy of continuous adaptation constitutes Incremental Learning (IL), with the goal of sequentially improving a model's overall performance. IL is a fundamental requirement for systems operating in dynamic environments that require sequential adaptation to new tasks, such as new operational domains, scenarios, environments, and situations, without erasing previously acquired capabilities, \ie, catastrophic forgetting~\cite{menezes_continual_2023, wang_comprehensive_2024}. However, edge hardware often lacks the computational power for on-device training, necessitating a paradigm in which updates are computed on a central server and transmitted to the edge device.

Reliable transmission must account for hardware and protocol constraints. The average 400 Byte minimum transport unit (MTU) for V2X applications using UDP or BTP necessitates fragmentation for larger data~\cite{postel1980user, molina-masegosa_empirical_2020, shimizu_comparison_2019}. Since these protocols lack automatic retransmission~\cite{postel1980user, ETSI2019BTP}, any lost fragment invalidates the update, requiring error correction or full retransmission~\cite{postel1980user, ma_approximate_2025, paolini_efficient_2024}. Consequently, the success probability drops rapidly with increasing update size (Figure~\ref{fig1}b). Furthermore, TCP Slow Start divides data into increasing windows~\cite{rfc9293}, each of which adds a round-trip time (RTT) unit for acknowledgments. However, if data fits the 14.6~KB initial congestion window using a handshake-free protocol, \mbox{0-RTT} transmission is achievable (\cf Figure~\ref{fig1}c)~\cite{rfc7413}.

To our knowledge, we are the first to address these communication constraints for IL. Limiting update sizes to 14.6~KB~\cite{rfc9293} within our dense coverage framework enables efficient model adaptation across TCP, UDP, and BTP hardware while generalizing to V2X, Wi-Fi, and 2G-5G technologies. Updates are server-trained and distributed to (potentially new) roadside units (RSUs), with transmission limited to vehicles entering the specific spatiotemporal context. We term this new framework Communication-Aware Incremental Learning (CAIL), which enables rapid performance improvements without adding vehicle-level computing or storage overhead.

Mixture-of-Experts (MoE) architectures offer a promising structural solution for this new paradigm by incorporating new knowledge into distinct modules~\cite{li_theory_2024}, scaling the system without overwriting a monolithic network. While this modularity resists catastrophic forgetting, it only solves the transmission challenge if the experts are sufficiently small. We introduce MECAIL (\textbf{M}o\textbf{E} for \textbf{C}ommunication-\textbf{A}ware \textbf{I}ncremental \textbf{L}earning), the first approach to improve performance using specialized experts small enough for efficient transfer. Treating the 14.6~KB window as a strict target, we employ an open-vocabulary object detection (OVOD) base model to efficiently learn new domains and classes. Each expert is $\sim$12~KB (16-bit) or $\sim$6~KB (8-bit), enabling transmission within a single TCP flight or minimal UDP/BTP messages to ensure reliability and efficiency even in high-speed, crowded environments.

We validate MECAIL on the challenging incremental object detection (IOD) benchmarks D-RICO~\cite{neuwirth-trapp_rico_2025-2} and ODinW-13~\cite{li_elevater_2022}. While our results match baselines on D-RICO, they highlight the challenges posed by the strict requirements on ODinW-13. Uniquely, MECAIL is the only approach that meets the aforementioned communication constraints.

Our contributions are as follows:

\begin{enumerate}
\item We introduce a dense expert coverage paradigm for spatiotemporal updates to specialize a monolithic network.
\item We formalize Communication-Aware Incremental Learning, limiting updates to 14.6~KB for efficient TCP and UDP/BTP transmission over V2X, Wi-Fi, and 2G-5G.
\item We present MECAIL as a lightweight approach ($\sim$12~KB per update) to sequentially increase a model's capability.
\end{enumerate}

\section{Related Works}\label{sec:rel-works}

\paragraph{Transmission-Constrained Incremental Learning}
IL mainly addresses sequential model updates while mitigating catastrophic forgetting~\cite{wang_comprehensive_2024,menezes_continual_2023}. Communication aspects are more extensively studied in federated learning~\cite{wen_survey_2023}, federated continual learning ~\cite{hamedi_federated_2025}, and distributed learning~\cite{ paolini_efficient_2024}. The two main approaches are either to make the communication more efficient by reducing or quantifying the transmission size~\cite{liu_fedet_2023, choudhary_codec_2023}, or by correcting errors and developing robust algorithms for lossy communication~\cite{paolini_efficient_2024}. We aim to enable TCP transmission over Wi-Fi and 2G-5G, and with V2X hardware, via connectionless UDP and BTP broadcasts. To achieve zero-round-trip-time (0-RTT) communication with TCP, protocols such as TCP Fast Open~\cite{rfc7413} and QUIC~\cite{rfc9000} are necessary, and the update size needs to fit within the 14.6 KB initial congestion window. V2X standards like ETSI ITS-G5~\cite{etsi302} and C-V2X~\cite{shimizu_comparison_2019, 3gpp2024UMTS} are optimized for small messages ($\sim$300--800~bytes). Research on sending larger data focuses mainly on automatic repeat request~\cite{bendrick_large_2024, bendrick_error_2024} or forward error correction~\cite{paolini_efficient_2024}. As errors accumulate exponentially, small update sizes are still favored. In this work, we focus on communication efficiency.

\paragraph{Incremental Object Detection}
Addressing catastrophic forgetting in IOD~\cite{wang_comprehensive_2024,menezes_continual_2023} primarily involves knowledge distillation, where student models mimic teachers to retain earlier capabilities~\cite{leonardis_bridge_2025}. Rehearsal methods offer an alternative solution by partially replaying prior tasks~\cite{monte_replay_2024}. Other directions include strengthening feature representations~\cite{mo_multi-level_2025} and designing specific optimization strategies~\cite{joseph_incremental_2022}. There is also a growing trend toward hybrid methods~\cite{ibrahim_node_2024}. An overview of prompt-based strategies for IOD is provided in~\cite{neuwirth-trapp_incremental_2025-2}. Unlike traditional closed-world IOD, our method is built for OVOD models and naturally accommodates new classes.

\paragraph{Incremental Open-Vocabulary Detection}\label{sec:rel-work:ilovod}
Dedicated solutions for OVOD remain scarce, however, using them for IOD has the unique benefit of naturally supporting new classes. While broad IL strategies like EWC~\cite{kirkpatrick_overcoming_2017}, GEM~\cite{lopez-paz_gradient_2017}, Distillation~\cite{li_continual_2025}, and Replay~\cite{wang_comprehensive_2024} are technically applicable, only two methods target this specific setting. ZiRa~\cite{deng_zero-shot_2025} relies on a frozen backbone and distills knowledge through dual fast and slow learning branches with reparameterization. Alternatively, DitHub~\cite{cappellino_dithub_2025} integrates class-specific modular adapters using LoRA. Recent studies emphasize that modality gaps between text and vision severely limit zero-shot inference and make it necessary to apply fine-tuning to concrete application domains~\cite{arandjelovic_three_2023-1}. MECAIL distinguishes itself from ZiRa and DitHub by prioritizing the reduction of transmitted parameters.

\paragraph{Mixture-of-Experts for IL} MoE frameworks mitigate catastrophic forgetting in IL by routing inputs to dedicated subnetworks, a paradigm established by Expert Gate~\cite{aljundi_expert_2017}. While subsequent Bayesian variants allowed adaptive expert growth~\cite{lee_neural_2019, ye_lifelong_2021}, current research prioritizes regulating expansion via expert consolidation~\cite{park_learning_2024}, parameter-efficient adapters~\cite{yu_boosting_2024}, and refined gating~\cite{le_mixture_2024-1}. Supported by theoretical scalability findings~\cite{li_theory_2024}, MECAIL leverages an MoE architecture to learn a small network that generates domain embeddings to alter class embeddings.
\begin{figure*}[t]
    \centering
    \includegraphics[width=0.70\linewidth, clip]{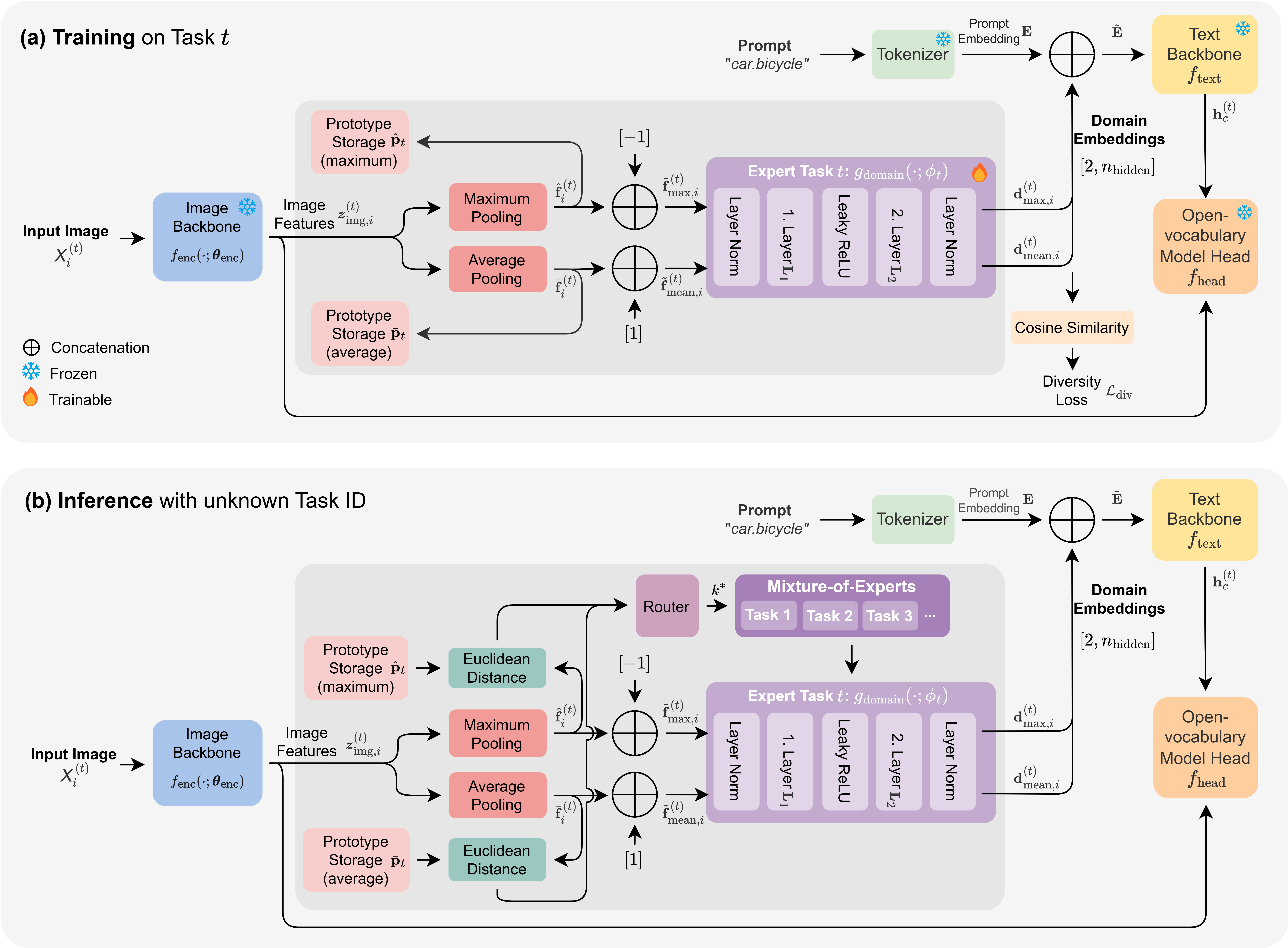}
\caption{\textbf{MECAIL method pipelines for Training and Inference.} \textbf{(a) Training} involves updating only the expert (two linear layers, Leaky ReLU, layer norms), which processes an image's global features to generate domain embeddings that condition the text backbone for prediction with a standard object detection loss for training and an additional diversity loss to encourage distinct embeddings. The pooled features are augmented by concatenating with $[1]$ and $[-1]$ to distinguish global average pooling (GAP) and global max pooling (GMP) features, and the domain and textual embeddings are combined without changing feature dimension. After training, the expert weights and static domain prototypes are saved. \textbf{(b) Inference} selects the expert based on the minimal distance between the max and average pooled features and their respective prototypes.}
    \label{fig:detail}
\end{figure*}

\section{Preliminaries}
\label{sec:preliminaries}

We set a transmission size of 14.6 KB for new tasks and domains as the design requirement to enable efficient sequential wireless updates. This allows us to fit the entire update into the initial congestion window of TCP~\cite{rfc9293} and results in a $\sim$50\% total success rate for V2X hardware with 0.4 KB packages and a 98\% success probability per package~\cite{molina-masegosa_empirical_2020, shimizu_comparison_2019}. This requirement can be met by reducing the number of parameters per update and increasing the quantization.

In this paper, we introduce the first method that satisfies this stringent constraint. Our objective is to progressively adapt an object detection model across a sequence of tasks~\cite{wang_comprehensive_2024}. We formalize the problem as a sequence of $T$ distinct tasks, represented by $\{\mathcal{T}_1, \dots, \mathcal{T}_T\}$. For every task $\mathcal{T}_t$, we utilize a dataset $\sD_t = \{(\tX_i^{(t)}, \sY_i^{(t)})\}_{i=1}^{n_t}$ comprising $n_t$ samples. Here, the input $\tX_i^{(t)} \in \R^{H \times W \times C}$ is an image tensor defined by its height $H$, width $W$, and channel count $C$. The associated ground truth $\sY_i^{(t)}$ contains a collection of $m_i^{(t)}$ object instances, formalized as $\sY_i^{(t)} = \{ (c_{i,j}^{(t)}, \vb_{i,j}^{(t)}) \}_{j=1}^{m_i^{(t)}}$. Within this set, each instance consists of its bounding box coordinates $\vb_{i,j}^{(t)} \in \R^4$ and a specific class $c_{i,j}^{(t)}$ belonging to the task-dependent class set $\sC_t$. Adhering to standard protocols for IL, the task identity is provided only during the training phase and is unavailable at inference~\cite{wang_comprehensive_2024}.

\section{Methodology}
\label{sec:methodology}

To enable sequential low-data updates, we adopt a MoE strategy in which a new expert is introduced and selected for each task and domain. We base our method on OVOD models that provide a strong foundation for minimizing the need to learn new information and for supporting efficient extension to arbitrary novel classes.

Our approach builds on the commonly used OVOD architecture~\cite{leonardis_grounding_2024}, which decomposes the model into three components. The image backbone $f_{\mathrm{img}}(\cdot; \vtheta_{\mathrm{img}})$ extracts visual features $\vz_{\mathrm{img},i}^{(t)}$ from an image $\tX_i^{(t)}$ and the text backbone $f_{\mathrm{text}}(\cdot; \vtheta_{\mathrm{text}})$ encodes prompts generated from the class set $\sC_t$, producing token-level embeddings. These embeddings are aggregated into a matrix of class-specific semantic embeddings $\mZ_{\mathrm{text}}^{(t)} \in \R^{L \times D}$, where $D$ is the text feature dimension and $L$ the number of tokens depending on the number of classes $|\sC_t|$. The detection head $f_{\mathrm{head}}(\cdot, \cdot; \vtheta_{\mathrm{head}})$ takes $\vz_{\mathrm{img},i}^{(t)}$ and $\mZ_{\mathrm{text}}^{(t)}$ as input to predict bounding boxes and class labels. The full parameter set for task $\mathcal{T}_t$ is denoted by $\vtheta_t = \{\vtheta_{\mathrm{img}}, \vtheta_{\mathrm{text}}, \vtheta_{\mathrm{head}}, \vtheta_{t,\mathrm{extra}}\}$, where $\vtheta_{t,\mathrm{extra}}$ contains the parameters introduced by each expert. The image and text backbones, as well as the detection head, are frozen and thus task-independent, eliminating the need for transmission.

MECAIL is motivated by parameter-efficient Textual Inversion~\cite{ruis_textual_2025}, which optimizes only the textual embeddings. However, learning distinct class embeddings per task captures inter-task but not intra-task variability. We therefore condition the class embeddings on image features and perform the adaptation via the text backbone, thereby avoiding an additional adaptation mechanism. The resulting design only needs to generate task- and image-specific vision embeddings from the image features, since the remainder is already part of the frozen OVOD model.

The architecture used for training and inference is shown in Figure~\ref{fig:detail}. During training, we train a lightweight two-layer MLP expert for each new task. Given global image features obtained via average and max pooling, the expert predicts two domain-specific embedding vectors. These are concatenated to the textual prompts before encoding, yielding visually grounded class embeddings for the detection head. In parallel, a static prototype is computed and stored for each task by aggregating the global image features over the task dataset. During inference, a test image is routed to the appropriate expert by selecting the nearest stored prototype in feature space. The selected expert generates the domain embeddings, which are prepended to the standard text prompts in the same manner as during training.

\subsection{Domain-Specific Embedding Generation}

For a given image $\tX_i^{(t)}$, we extract features using both Global Average Pooling (GAP) and Global Max Pooling (GMP), yielding the vectors $\bar{\vf}_i^{(t)}$ and $\hat{\vf}_i^{(t)} \in \R^{d_{\text{feat}}}$, respectively, where $d_{\text{feat}}$ is the dimension of the image feature space. This extracts two different kinds of information from the image: GAP captures the average activation strength of each feature across the spatial domain, and GMP captures its maximal activation, thereby characterizing overall presence versus peak presence while discarding spatial localization~\cite{lin_network_2014}.  To enable the expert to learn distinct transformations for each pooling type, we then concatenate a scalar type indicator ($+1$ for GAP and $-1$ for GMP) to each feature vector. This process creates the augmented input vectors    $\tilde{\vf}_{\text{mean}, i}^{(t)} = [\bar{\vf}_i^{(t)}; 1]_f \in \R^{d_{\text{feat}}+1}$ and $\tilde{\vf}_{\text{max}, i}^{(t)} = [\hat{\vf}_i^{(t)}; -1]_f \in \R^{d_{\text{feat}}+1}$. Here, $[\cdot;\cdot]_f$ means concatenation feature-wise, \ie, the -1 and 1 are added as an additional feature dimension.

For each task $\mathcal{T}_t$, we process these augmented vectors with a dedicated expert $g_{\text{domain}}(\cdot; \vphi_t)$, with trainable parameters $\vphi_t$. The expert consists of: (1) an input Layer Normalization to mitigate covariate shifts across domains, (2) a small MLP with two linear layers, and (3) a final Layer Normalization to align the statistical properties of the resulting embeddings with our text embeddings. Concretely, the MLP part includes a bottleneck layer $\mL_1: \R^{d_{\text{feat}}+1} \to \R^{d_{\mathrm{lat}}}$ and an expansion layer $\mL_2: \R^{d_{\mathrm{lat}}} \to \R^D$, where  $d_{\mathrm{lat}}$ is the dimension of the latent space, and $D$ that of the text embedding. The choice of $d_{\mathrm{lat}}$ is crucial, as it influences both expert capacity and efficiency. A non-linear activation function $\sigma$ is applied between the two linear layers. Thus, the overall mapping is
\begin{equation}
    g_{\text{domain}}(\cdot; \vphi_t) = \mathrm{LayerNorm}(\mL_2(\sigma(\mL_1(\mathrm{LayerNorm}(\cdot))))).
\end{equation}

The final domain embeddings $\vd_{\text{mean},i}^{(t)}, \vd_{\text{max},i}^{(t)} \in \R^D$, which form the conditional signal for the text encoder, are
\begin{align}
\label{eq:output_embeddings_mean}
    \vd_{\text{mean},i}^{(t)} &= g_{\text{domain}}(\tilde{\vf}_{\text{mean}, i}^{(t)}; \vphi_t) \\
\label{eq:output_embeddings_max}
    \vd_{\text{max},i}^{(t)} &= g_{\text{domain}}(\tilde{\vf}_{\text{max}, i}^{(t)}; \vphi_t)
\end{align}

\subsection{Embedding Concatenation and Text Encoding}

The text prompts for the class set $\sC_t$ are tokenized into sequences of $L$ token embeddings $\mE = [\ve_{1}, \ldots, \ve_{L}] \in \R^{L \times D}$, where each $\ve_{j} \in \R^D$ corresponds to the embedding of the $j$-th token. To condition these prompts on the visual domain, we append the two domain embeddings from \eqref{eq:output_embeddings_mean} and \eqref{eq:output_embeddings_max}, yielding
\begin{equation}
    \tilde{\mE} = [\mE, \vd_{\text{mean},i}^{(t)}, \vd_{\text{max},i}^{(t)}]_v \in \R^{(L+2) \times D},
\end{equation}
where $[\cdot,\cdot]_v$ denotes concatenation along the sequence dimension, preserving the embedding dimension $D$ while increasing the sequence length. The augmented sequence $\tilde{\mE}$ is then passed to the text backbone $f_{\text{text}}$, to enable interactions between the original tokens and the appended domain embeddings, \ie, $\vh^{(t)} = f_{\text{text}}(\tilde{\mE})\in \R^{(L+2) \times D}$. For the final detection and classification tasks, the two appended domain-specific embeddings are removed from this sequence, resulting in the final class embeddings $\vh^{(t)}_c \in \R^{L \times D}$. We modify the attention mask to enable bidirectional attention between domain embeddings and class tokens. To differentiate these components in the shared space, domain-embedding positional encodings are offset by the maximum class-token index.

\subsection{Training}
During the training phase for a new task $\mathcal{T}_t$, the parameters of the image backbone ($\vtheta_{\text{img}}$), text backbone ($\vtheta_{\text{text}}$), and detection head ($\vtheta_{\text{head}}$) are frozen. Training is highly efficient, as we only optimize the parameters $\vphi_t$ of the newly introduced expert MLP. The total loss function is
\begin{equation}
    \mathcal{L}^{(t)} = \mathcal{L}_{\text{det}}(\tX^{(t)}, \sY^{(t)}; \vphi_t) + \lambda_{\mathrm{div}}\mathcal{L}_{\text{div}}(\vphi_t),
\end{equation}
with $\mathcal{L}_{\text{det}}$ being sum of the standard object detection losses~\cite{leonardis_grounding_2024}. The diversity loss, $\mathcal{L}_{\text{div}}$ and its respective scaling $\lambda_{\mathrm{div}}$, encourages the two generated domain embeddings to be decorrelated. This pushes the expert to learn complementary, non-redundant representations from the GAP and GMP features. This instance-level loss is formulated as the expected cosine similarity to enforce orthogonality.
\begin{equation}
\mathcal{L}_{\text{div}}(\phi_t) = \mathbb{E}_{i \in \mathcal{D}_t} \left[ \left( \frac{\mathbf{d}_{\text{mean},i}^{(t)} \cdot \mathbf{d}_{\text{max},i}^{(t)}}{\|\mathbf{d}_{\text{mean},i}^{(t)}\|_2 \|\mathbf{d}_{\text{max},i}^{(t)}\|_2} \right)^2 \right].
\end{equation}

\subsection{Expert Routing}~\label{sec:routing}

Following \cite{wang_non-exemplar_2024}, we employ an optimization-free routing based on image features. During training, we aggregate GMP and GAP vectors across the task dataset to compute mean domain prototypes $\bar{\vp}_t = \frac{1}{n_t} \sum_{i=1}^{n_t} \bar{\vf}_i^{(t)}$ and $\hat{\vp}_t = \frac{1}{n_t} \sum_{i=1}^{n_t} \hat{\vf}_i^{(t)}$, which serve as condensed representations of domain $\mathcal{T}_t$. For inference, we select the expert $k^*$ by identifying the stored prototype with the smallest combined distance to the input's GMP and GAP features.

\begin{equation}
    k^* = \underset{k \in \{1, \dots, T\}}{\arg\min} \left(  \|\bar{\vf} - \bar{\vp}_k\|_2^2 + \|\hat{\vf} - \hat{\vp}_k\|_2^2\right)
\end{equation}

\subsection{Inference}\label{sec:inference}

At inference, the task identity of a test image $\tX$ is unknown, and an expert MLP $g_{\text{domain}}(\cdot; \vphi_{k^*})$ is selected via the routing described in Section~\ref{sec:routing}. The expert generates the domain embeddings from the GAP and GMP features, which are concatenated to the task-specific prompt embedding and processed by the text encoder. When applying MECAIL, if the task/domain is known during inference, the prototypes and the associated nearest-mean classifier can be omitted.
\begin{table*}[t]
\centering
\small
\caption{\textbf{Main results on transmission-efficiency, D-RICO and ODinW-13.} "Avg" is the mean precision across tasks (15 for D-RICO, 13 for ODinW). "Param." is the number of transmitted weights and "Size" the corresponding 16-bit update size in KB. For TCP, \#Seg. counts 1,460-byte segments and \#RTTs the round-trips to deliver them under Slow Start. For V2X, \#Chunks counts 400-byte chunks with a success probability of 98\% per chunk. MECAIL approaches baseline performance at substantially higher communication efficiency. Best in bold.}
\label{tab:combined-results}
\renewcommand{\arraystretch}{0.93}
\setlength{\tabcolsep}{2pt}
\begin{tabular}{lrr|rc|rc|cc|cccc}
\toprule
 & & & \multicolumn{2}{c|}{TCP} & \multicolumn{2}{c|}{V2X} & \multicolumn{2}{c|}{\textbf{D-RICO (Avg)}} & \multicolumn{4}{c}{\textbf{ODinW-13 (Avg)}} \\
\textbf{Method} & \textbf{Param.} $\downarrow$ & \textbf{Size} $\downarrow$ & \textbf{\#Seg.} $\downarrow$ & \textbf{\#RTTs} $\downarrow$ & \textbf{\#Chunks} $\downarrow$ & \textbf{Succ.} $\uparrow$ & \textbf{1-shot} $\uparrow$ & \textbf{Full} $\uparrow$ & \textbf{1-shot} $\uparrow$ & \textbf{5-shot} $\uparrow$ & \textbf{10-shot} $\uparrow$ & \textbf{Full} $\uparrow$ \\
\midrule
Zero-shot GDINO &-&-&-&-&-&-&\multicolumn{2}{c|}{17.43}&\multicolumn{4}{c}{46.80}\\
\midrule
TFA~\cite{wang_frustratingly_2020} & 397k & 775 & 544 & 5 & 1985 & <0.0001 & 14.76 & 27.08 & 39.50 & 45.76 & 46.61 & 47.93 \\
iDETR~\cite{dong_incremental-detr_2023} & 2117k & 4134 & 2900 & 8 & 10584 & <0.0001 & \textbf{26.70} & \textbf{31.58} & 49.82 & 51.65 & 53.29 & 58.71 \\
AT~\cite{houlsby_parameter-efficient_2019} & 412k & 806 & 565 & 5 & 2063 & <0.0001 & 26.07 & 31.47 & 46.23 & 47.16 & 47.34 & 51.14 \\
ZiRa~\cite{deng_zero-shot_2025} & 2311k & 4515 & 3167 & 8 & 11558 & <0.0001 & 26.61 & 30.62 & \textbf{50.20} & \textbf{54.19} & \textbf{54.86} & 59.73 \\
DitHub~\cite{cappellino_dithub_2025} & 1505k & 2940 & 2063 & 7 & 7527 & <0.0001 & - & 30.37 & 49.19 & 52.85 & 54.43 & \textbf{62.19} \\
MECAIL (ours) & \textbf{6k} & \textbf{12} & \textbf{9} & \textbf{0} & \textbf{30} & \textbf{0.55} & 25.35 & 30.65 & 48.27 & 49.63 & 48.03 & 52.69 \\
\bottomrule
\end{tabular}
\vspace{-0.5cm}
\end{table*}

\section{Experiments}\label{sec:exps}

\subsection{Setup}\label{sec:setup}

\paragraph{Implementation and Model}
We follow the experimental setup of \cite{deng_zero-shot_2025,cappellino_dithub_2025} and evaluate MECAIL on Grounding DINO (GDINO)~\cite{leonardis_grounding_2024} with a Swin-Tiny~\cite{liu_swin_2021} image backbone and a BERT~\cite{devlin_bert_2019} text backbone. The base model is kept frozen, and MECAIL introduces one task-specific expert trained for each incremental task. Training is performed over 2 epochs per task, with standard image augmentation and a batch size of 4. Further implementation and training details are provided in the Appendix.

\paragraph{Evaluation Metric}
We report results using the COCO evaluation protocol, where $\mAP=\mathrm{mAP}@\text{[.5:.05:.95]}$~\cite{fleet_microsoft_2014}. Let $\mAP_{k,j}$ denote the $\mAP$ obtained on task $\mathcal{T}_j$ after training up to task $\mathcal{T}_k$, with $j \leq k$. Overall performance after learning all $T$ tasks is measured by the mean over tasks, defined as $\mathrm{Avg} = \tfrac{1}{T}\sum_{j=1}^{T}\mAP_{T,j}$~\cite{deng_zero-shot_2025}.

\paragraph{Benchmarks} We evaluate MECAIL on D-RICO~\cite{neuwirth-trapp_rico_2025-2} and ODinW-13~\cite{li_elevater_2022}. D-RICO assesses robustness to domain shifts in driving and surveillance data, whereas ODinW-13 evaluates adaptation to novel semantic domains with disjoint labels. Details are in the Appendix. Following \cite{deng_zero-shot_2025}, we vary shot settings (\ie, training samples per task).

\paragraph{Baselines}
For D-RICO, we compare MECAIL with TFA~\cite{wang_frustratingly_2020},  iDETR~\cite{dong_incremental-detr_2023}, AT~\cite{houlsby_parameter-efficient_2019}, ZiRa~\cite{deng_zero-shot_2025}, and DitHub~\cite{cappellino_dithub_2025}. Baseline selection and results for ODinW-13 are taken from \cite{deng_zero-shot_2025} \cite{cappellino_dithub_2025} and for D-RICO obtained by us. We also provide GDINO's zero-shot performance as a reference.

\subsection{Results}

\paragraph{Communication Efficiency}
Table~\ref{tab:combined-results} presents the communication efficiency results. MECAIL requires transmitting $\sim$6k parameters, significantly fewer than other methods, resulting in only 12KB of 16-bit data. This reduction implies fewer TCP segments and UDP/BTP chunks, resulting in 0-RTT for TCP and a 55\% total success probability for UDP/BTP. Conversely, other approaches need a number of RTTs between 5 and 8 to deliver the update and exhibit a practically zero UDP/BTP success probability. Therefore, MECAIL significantly outperforms the alternatives in terms of communication efficiency.

\paragraph{D-RICO}
Table~\ref{tab:combined-results} further reports the results on D-RICO. The zero-shot performance of GDINO is markedly lower here (17.43 mAP) than that of ODinW-13 (46.80 mAP), indicating that the pre-trained model encodes less knowledge of the D-RICO domains. MECAIL delivers substantial gains in both the 1-shot and full-training settings. While competing methods achieve comparable performance with MECAIL, the latter is substantially more communication-efficient.

\paragraph{ODinW-13}
Additionally, Table~\ref{tab:combined-results} presents MECAIL's results compared to prior approaches on ODinW-13. MECAIL consistently outperforms the zero-shot GDINO baseline across all shot settings, demonstrating its ability to adapt from limited data. In the 1-shot regime, MECAIL matches the performance of parameter-intensive alternatives and scales favorably with additional training data. However, the other approaches profit more strongly from additional data. DitHub achieves the highest overall average mAP across full training, while ZiRa achieves the highest mAP in the few-shot settings, albeit at the cost of substantially greater parameter overhead.

\begin{table}
    \centering
    \footnotesize
    \renewcommand{\arraystretch}{1}
    \setlength{\tabcolsep}{1pt}
    \caption{\textbf{Ablation on latent dimension $d_\mathrm{lat}$ and learning rate (lr)}. A latent dimension of $d_\mathrm{lat}=10$ outperforms 5 slightly, but does not fit in the initial congestion window of TCP, and V2X success is lower. The optimal learning rate depends on the latent dimension. Zero-shot GDINO performance is 46.80. Metrics and column definitions follow Table~\ref{tab:combined-results}. Best in bold.}
    \label{tab:ablation-latent-dim-lr}
    \renewcommand{\arraystretch}{0.93}
    \setlength{\tabcolsep}{0.4pt}
    \begin{tabular}{lcc|cc|c|c|cc}
        \toprule
        & & & & & \multicolumn{1}{c|}{TCP} & \multicolumn{1}{c|}{V2X} & \multicolumn{2}{c}{\textbf{ODinW (Avg)}} \\
\textbf{Method} & $d_\mathrm{lat}$ & lr & \textbf{Param}$\downarrow$ & \textbf{Size}$\downarrow$ & \textbf{\#RTTs}$\downarrow$ & \textbf{Succ.}$\uparrow$ & \textbf{1-shot}$\uparrow$ & \textbf{Full}$\uparrow$ \\
        \midrule
        ZiRa~\cite{deng_zero-shot_2025} & - & - & 2311k & 4515 & 9 & <0.0001 & 48.56 & 57.98 \\
        DitHub~\cite{cappellino_dithub_2025} & - & - & 1505k & 2940 & 8 & <0.0001 & \textbf{49.19} & \textbf{62.19} \\
        \midrule
        MECAIL (ours) & 5 & 0.005 & \textbf{6k} & \textbf{12} & \textbf{0} & \textbf{0.55} & 47.77 & 51.07 \\
        MECAIL (ours) & 5 & 0.001 & \textbf{6k} & \textbf{12} & \textbf{0} & \textbf{0.55} & 48.38 & 52.43 \\
        MECAIL (ours) & 5 & 0.0005 & \textbf{6k} & \textbf{12} & \textbf{0} & \textbf{0.55} & 48.05 & 52.72 \\
        MECAIL (ours) & 10 & 0.005 & 12k & 23 & 1 & 0.30 & 48.69 & \textbf{53.01} \\
        MECAIL (ours) & 10 & 0.001 & 12k & 23 & 1 & 0.30 & \textbf{48.74} & 52.68 \\
        MECAIL (ours) & 10 & 0.0005 & 12k & 23 & 1 & 0.30 & 48.06 & 52.06 \\
        MECAIL (ours) & 20 & 0.005 & 24k & 46 & 2 & 0.09 & 46.06 & 51.11 \\
        MECAIL (ours) & 20 & 0.001 & 24k & 46 & 2 & 0.09 & 46.50 & 52.67 \\
        MECAIL (ours) & 20 & 0.0005 & 24k & 46 & 2 & 0.09 & 46.86 & 52.60 \\
        \bottomrule
    \end{tabular}

    \vspace{10pt}
    
\caption{\textbf{Ablation on visual conditioning and diversity loss.} Evaluation here is on a subset of ODinW-13 (\cf, Suppl. Mat.) with a zero-shot GDINO result of 45.74. Best in bold.}
    \label{tab:ablation-part1}
    \renewcommand{\arraystretch}{0.93}
    \begin{tabular}{lcccc}
        \toprule
        & & & \multicolumn{2}{c}{\textbf{ODinW (Avg)}} \\
        \textbf{Method} & \textbf{Setting} & & \textbf{1-shot} $\uparrow$ & \textbf{Full} $\uparrow$ \\
        \midrule
        MECAIL (ours)    & GAP \& GMP cond. & & \textbf{45.26} & \textbf{53.50} \\ 
        MECAIL (ours)    & only GAP cond. & & 39.93 & 50.88 \\
        MECAIL (ours)    & only GMP cond. & & 38.93 & 52.66 \\
        MECAIL (ours)    & w/o vis. cond. & & 38.43 & 45.89 \\
        \midrule
        MECAIL (ours)    & $\lambda_\mathrm{div}=1$ & & \textbf{45.26} & \textbf{53.50} \\
        MECAIL (ours)    & $\lambda_\mathrm{div}=0$ & & 40.73 & 53.45 \\
        \bottomrule
    \end{tabular}

    \vspace{10pt}

   \caption{\textbf{Ablation on routing and task-incremental learning (TIL).} For TIL, the task ID is known, showing the model's capability. Best in bold.}
    \label{tab:ablation-part2}
    
    \renewcommand{\arraystretch}{0.93}
    \begin{tabular}{lcccc}
        \toprule
        & & & \multicolumn{2}{c}{\textbf{ODinW (Avg)}} \\
        \textbf{Method} & \textbf{Setting} & & \textbf{1-shot} $\uparrow$ & \textbf{Full} $\uparrow$ \\
        \midrule
        MECAIL (ours)    & GMP \& GAP Router & & \textbf{48.27} & \textbf{52.69} \\
        MECAIL (ours)    & GMP Router & & 47.90 & 51.18 \\
        MECAIL (ours)    & GAP Router & & 47.28 & 51.57 \\
        \midrule
        MECAIL (ours)    & Task ID unknown & & 48.27 & 52.69 \\
        MECAIL (ours)    & Task ID known & & \textbf{49.08} & \textbf{55.81} \\
        \bottomrule
    \end{tabular}
    \vspace{-15pt}
\end{table}

\subsection{Ablation Study}\label{sec:ablation}
\paragraph{Latent Dimension and Learning Rate}
MECAIL introduces a few hyperparameters, with the central one being the latent dimension $d_\mathrm{lat}$ of the expert $g_{\text{domain}}(\cdot; \vphi_t)$ and the learning rate (lr). Table~\ref{tab:ablation-latent-dim-lr} shows that MECAIL is robust to both. Increasing $d_\mathrm{lat}$ from 5 to 10 yields only minor gains, and 20 offers no further benefit. However, 10 and 20 both exceed the requirement of an update size below 14.6 KB.

\paragraph{Maximum and Average Pooling Duality}
MECAIL conditions the text prompt embedding with both GMP and GAP. Table~\ref{tab:ablation-part1} shows their combination is best, particularly for 1-shot, while either alone is weaker. GMP outperforms GAP in full training, but the reverse holds for 1-shot. Removing visual conditioning, i.e.\ learning the domain embeddings directly, degrades performance the most. These experiments used only a subset of the tasks.

\paragraph{Diversity Loss}
We use a diversity loss to encourage distinct GAP- and GMP-based embeddings. Table~\ref{tab:ablation-part1} shows that it improves 1-shot performance, while under full training, its effect is negligible. This indicates its value mainly under data-scarce conditions.

\paragraph{Task Incremental Learning and Routing}
MECAIL routes tasks using both GAP and GMP features with their prototypes. Table~\ref{tab:ablation-part2} shows that combining them outperforms using either alone. In task incremental learning (TIL), where task IDs are known at inference, MECAIL achieves the highest performance, reflecting its representational capacity.

\section{Discussion}

\paragraph{On Real-world Application}
The experimental setting differs from deployment but demonstrates the method’s potential. MECAIL consistently improves zero-shot performance, indicating that tiny experts enable adaptation. In practice, the method assumes a strong base model that is minimally adapted to a target domain or task. \smalltriangleright \textit{Real-world applications of MECAIL depend on a solid base model.}

\paragraph{On Quantization} Communication costs depend on data size, not parameter count, allowing the addition of parameters when precision is reduced (8 or 4 bits). However, 
low-precision quantization typically requires pre-deployment fine-tuning. \smalltriangleright \textit{Quantization is a central parameter for CAIL.}

\paragraph{On the 14.6 KB Requirement}
While the 14.6 KB target reflects common TCP segmentation, real systems vary. V2X hardware transmits sequential $\sim$0.4~KB segments without an initial congestion window and exhibits a lower per-packet success probability, thereby favoring small updates~\cite{molina-masegosa_empirical_2020, shimizu_comparison_2019}. \smalltriangleright \textit{The update-size target depends on hardware \& requirements.}

\paragraph{On the Full-training Discrepancy} While MECAIL matches higher-parameter methods on D-RICO, it lags on ODinW-13 under full training. This gap stems from lower base-model performance, greater task diversity, and D-RICO's smaller class count. \smalltriangleright \textit{MECAIL entails an efficiency-performance tradeoff.}

\paragraph{On the Parameter Count} While efficiency often implies minimizing new parameters, methods like EWC and Replay still require transmitting all updated parameters. We argue update data size is the relevant metric in communication-constrained settings and propose a communication-aware approach. \smalltriangleright \textit{The definition of efficiency is context-dependent.}

\paragraph{On Feasibility} The feasibility of MECAIL is supported by the projected evolution of V2X communication. While current systems prioritize low-bandwidth signals, future specifications are designed to support raw sensor sharing and larger data packets~\cite{delooz_design_2024}. Consequently, the barriers to the transmission of experts are temporary, with upcoming standards paving the way for full implementation. \smalltriangleright \textit{Evolving V2X capabilities will enable real-world application.}

\section{Conclusion}

In this work, we demonstrated that we can substantially reduce the model update size required to learn new environments and scenarios sequentially. We achieved an update size below 14.6 KB, enabling direct TCP communication over Wi-Fi and 2G to 5G within the initial congestion window, and increasing transmission success probability for V2X hardware via UDP/BPT by maintaining low fragmentation. These results enable dense spatiotemporal coverage through experts that adapt a monolithic network, with spatiotemporal or feature-based expert selection using task prototypes. This establishes a new paradigm for intelligent transportation systems with millions of dedicated experts and rapid updates for hazard hotspots and novel environments, without large-scale retraining and validation.

\clearpage

\begin{center}
    \normalfont\normalsize\textbf{Appendix}
\end{center}
\vspace{0em} 

\setcounter{section}{0}
\renewcommand{\thesection}{\Alph{section}}

\makeatletter
\renewcommand\section{\@startsection {section}{1}{\z@}%
                                   {-1.5ex \@plus -0.5ex \@minus -.2ex}%
                                   {1.0ex \@plus .2ex}%
                                   {\normalfont\normalsize\itshape}}

\renewcommand{\@seccntformat}[1]{%
  \ifnum\pdfstrcmp{#1}{section}=0
    Appendix \csname the#1\endcsname.\hspace{0.5em}%
  \else
    \csname the#1\endcsname\quad
  \fi
}
\makeatother

\setcounter{figure}{0}
\setcounter{table}{0}
\renewcommand{\thefigure}{\thesection.\arabic{figure}}
\renewcommand{\thetable}{\thesection.\arabic{table}}

\section{Detailed Results}
Table~\ref{tab:main-rico-suppl} and~\ref{tab:main-odinw-suppl} show detailed results on D-RICO and ODinW-13.

\begin{table*}[b]
\centering
\tiny
\setlength{\tabcolsep}{4pt}
\renewcommand{\arraystretch}{0.60}
\caption{\textbf{Detailed results on D-RICO.} Best in bold. Dataset abbreviations in Appendix C.}
\label{tab:main-rico-suppl}
\resizebox{\textwidth}{!}{
\begin{tabular}{clccccccccccccccc>{\columncolor{lightpruple}}c}
\toprule
Shots & Method  & daytm & therm & fyfix & drone & simul & fycar & rgbth & vgame & night & fyind & gated & prsim & tfyin & incle & evcam & \textbf{Avg} \\
\midrule
0 & Grounding DINO & 28.40&28.40&17.88& 6.92&16.00&18.40&29.68&15.98&37.06&12.12&15.99& 16.33&19.89&0.00&9.32 & 17.43 \\
\midrule
\multirow{5}{*}{1} & TFA & 16.20&12.34&4.51&10.01&9.99&15.47&8.14&38.26&7.04&9.56&8.65&13.42&38.98&20.62&8.18 & 14.76 \\
 & iDETR & 30.73&24.63&10.55&17.70&20.87&\textbf{36.81}&\textbf{23.76}&45.04&15.29&19.15&19.55&\textbf{24.66}&\textbf{58.29}&41.58&11.83 & \textbf{26.70} \\
 & AT & 32.29&\textbf{24.67}&8.72&\textbf{18.74}&\textbf{21.66}&36.50&22.60&\textbf{49.74}&14.60&13.68&19.85&24.06&52.49&41.19&10.26 & 26.07 \\
 & ZiRa & 26.68&23.49&\textbf{10.70}&16.89&21.32&33.41&23.14&40.36&12.55&19.17&18.22&23.17&57.86&39.14&\textbf{15.08} & 26.61 \\
 & MECAIL (ours) & \textbf{33.02}&21.86&10.45&14.68&17.76&31.05&22.44&48.66&\textbf{16.94}&\textbf{22.22}&\textbf{20.48}&21.74&45.52&\textbf{42.92}&10.60 & 25.35 \\
\midrule
\multirow{7}{*}{Full} & TFA & 41.51&24.42&9.98&16.19&19.65&31.72&28.41&38.46&19.88&19.68&24.51&21.61&46.35&50.90&12.99 & 27.08 \\
& iDETR& 45.22&29.10&\textbf{12.77}&16.94&\textbf{25.90}&36.80&32.54&48.25&\textbf{23.18}& 22.52&28.10&25.35&48.66&56.67&21.71 & \textbf{31.58} \\
& AT & \textbf{46.04}&\textbf{30.28}&10.03&16.39&25.63&32.48&\textbf{38.31}&42.05&22.55&20.92&\textbf{30.64}&23.50&52.05&\textbf{59.20}&\textbf{21.99} & 31.47 \\
& ZiRa & 44.97&28.37&11.61&17.36&23.54&36.50&33.20&47.15&22.40&22.13&27.73&24.62&32.46&56.51&19.56 & 30.62 \\
& MECAIL (ours) & 36.32&27.50&11.80&\textbf{19.61}&21.87&\textbf{39.33}&32.87&\textbf{51.77}&20.04&\textbf{26.16}&22.17&\textbf{25.39}&\textbf{64.76}&48.07&12.06 & 30.65 \\
\bottomrule
\end{tabular}}
\end{table*}

\begin{table*}[b]
\centering
\renewcommand{\arraystretch}{0.60}
\tiny
\caption{\textbf{Detailed results on ODinW-13.} Best in bold. Dataset abbreviations in Appendix C.}
\label{tab:main-odinw-suppl}
\resizebox{\textwidth}{!}{
\begin{tabular}{clccccccccccccc>{\columncolor{lightpruple}}c}
\toprule
Shots & Method & Ae & Aq & Co & Eg & Mu & Pa & Pv & Pi & Po & Ra & Sh & Th & Ve & \textbf{Avg} \\
\midrule
0 & Grounding DINO & 19.11 & 20.82 & 64.75 & 59.98 & 25.34 & 56.27 & 54.80 & 65.94 & 22.13 & 62.02 & 32.85 & 70.38 & 57.07 & 46.80 \\
\midrule
\multirow{5}{*}{1} & TFA & 18.25 & 15.81 & 63.90 & 50.79 & 28.47 & 50.37 & 29.49 & 59.16 & 21.90 & 50.67 & 19.86 & 60.85 & 43.97 & 39.50 \\
 & iDETR & \textbf{22.81} & 23.24 & 69.75 & 61.43 & 31.73 & 56.27 & 55.40 & 62.44 & 28.45 & 60.33 & \textbf{43.33} & 73.64 & 58.84 & 49.82 \\
 & AT & 21.55 & 23.62 & 66.60 & 58.96 & 27.68 & 53.97 & 54.58 & 62.47 & 26.94 & 53.17 & 20.37 & 70.31 & 60.71 & 46.23 \\
 & ZiRa & 19.91 & \textbf{24.84} & 68.82 & 63.02 & 37.02 & 60.20 & 55.11 & 63.37 & 28.26 & \textbf{66.35} & 38.97 & 69.27 & 57.43 & 50.20 \\
 &DitHub & 21.33 & 24.34 & 69.23 & 61.99 & 42.10 & 57.08 & \textbf{58.85} & 55.24 & 24.62 & 57.58 & 33.56 & 70.99 & \textbf{62.50} & 49.19 \\
 & MECAIL (ours) &20.67&17.66&63.64&56.40&36.85&\textbf{61.68}&52.80&\textbf{66.04}&15.24&65.54&34.22&\textbf{79.02}&57.80&48.27 \\
  & MECAIL-TIL (ours) &20.67&19.77&69.82&61.45&37.885&54.68&55.980&63.21&20.01&64.31&33.46&\textbf{79.02}&57.81&49.08 \\

\midrule
\multirow{5}{*}{5} & TFA & 21.92 & 22.30 & 67.40 & 60.72 & 30.63 & 53.56 & 46.80 & 63.60 & 26.88 & 56.26 & 28.00 & 64.28 & 52.49 & 45.76 \\
& iDETR & 25.69 & 25.53 & 70.42 & 62.98 & 49.98 & 50.54 & 54.85 & 64.80 & 33.24 & 57.64 & \textbf{42.36} & 76.51 & 56.92 & 51.65 \\
 & AT& 14.63 & 24.97 & 66.56 & 64.19 & 38.85 & 42.03 & 55.49 & 65.20 & 27.48 & 52.68 & 32.21 & 71.23 & 57.54 & 47.16 \\
 & ZiRa& 24.13 & \textbf{30.92} & \textbf{72.46} & \textbf{66.34} & \textbf{51.23} & \textbf{56.27} & 60.35 & \textbf{67.44} & \textbf{36.22} & 60.18 & 42.22 & 77.18 & 59.55 & \textbf{54.19} \\
 &DitHub & \textbf{26.84} & 28.91 & 68.46 & 60.25 & 51.93 & 55.10 & \textbf{60.43} & 59.44 & 33.18 & \textbf{68.12} & 38.05 & 75.04 & \textbf{61.34} & 52.85 \\
 & MECAIL (ours) &20.26&22.37&68.89&62.35&45.19&55.38&49.97&65.39&27.32&61.74&31.23&\textbf{78.31}&56.84&49.63 \\
\midrule
\multirow{5}{*}{10} & TFA & 21.17 & 22.16 & 66.82 & 60.63 & 32.35 & 50.15 & 55.54 & 64.98 & 27.59 & 57.40 & 28.14 & 66.82 & 52.11 & 46.61 \\
& iDETR & 25.39 & 27.70 & 65.62 & 67.58 & 47.99 & 60.20 & 56.32 & 63.93 & 35.18 & 59.37 & \textbf{53.63}& 74.70 & 55.12 & 53.29 \\
 & AT & 18.73 & 25.42 & \textbf{69.77} & 66.34 & 35.84 & 48.25 & 53.39 & 64.07 & 28.89 & 50.33 & 32.50 & 66.57 & 55.37 & 47.34 \\
 & ZiRa & 25.01 & 29.44 & 69.59 & \textbf{68.44} & \textbf{54.72} & \textbf{61.68} & 59.69 & \textbf{67.14} & 36.05 & 61.77 & 48.30 & 73.92 & 57.47 & \textbf{54.86} \\
 &DitHub& \textbf{26.19} & \textbf{30.28} & 68.11 & 65.73 & 52.82 & 56.82 & \textbf{63.36} & 61.85 & \textbf{36.68} & \textbf{66.82} & 41.46 & \textbf{76.00} & \textbf{61.48} & 54.43 \\
 & MECAIL (ours) &24.40&21.18&64.56&61.57&40.47&55.40&49.43&63.64&26.92&60.61&25.77&75.42&55.05&48.03 \\
\midrule
\multirow{7}{*}{Full} & TFA & 23.80 & 30.65 & 67.21 & 61.77 & 30.52 & 50.23 & 47.73 & 60.91 & 29.25 & 61.72 & 31.42 & 66.23 & 61.61 & 47.93 \\
& iDETR& 32.64 & 46.65 & 70.99 & 68.56 & 55.32 & 58.88 & 64.48 & 71.01 & 50.33 & 63.30 & 39.19 & 77.12 & 64.80 & 58.71 \\
& AT & 23.62 & 39.90 & 72.32 & 65.51 & 31.47 & 50.48 & 60.51 & 66.07 & 39.09 & 53.50 & 34.04 & 68.07 & 60.23 & 51.14 \\
& ZiRa & 32.81 & 48.19 & 70.33 & 69.67 & \textbf{59.33} & 58.05 & 64.04 & 70.67 & 50.06& 67.49 & 45.51 & 76.76 & 63.54 & 59.73 \\
&DitHub& \textbf{34.62} & \textbf{50.65} & 70.46 & 68.56 & 49.28 & 
\textbf{65.57} & \textbf{69.58} & \textbf{71.10} & \textbf{56.65} & 70.88 & \textbf{52.82} & \textbf{79.30} & \textbf{68.18} & \textbf{62.19} \\
& MECAIL (ours) &27.68&28.05& 68.05& 57.57 &52.87  &59.21&42.09&67.64&36.41&73.39&36.57 &77.93&57.47&52.69 \\
& MECAIL-TIL (ours) &27.51&36.60& 68.37& 66.47 &43.35  &56.83&63.79&68.31&38.46&73.53&45.06 &78.27&58.97&55.81 \\
\bottomrule
\end{tabular}}

\end{table*}

\section{Details on the Implementation}\label{sec:reproduce}
We build upon the ZiRa repository~\cite{deng_zero-shot_2025}, adopting its training pipeline and structure. Each expert uses two linear layers: the first applies Kaiming uniform initialization~\cite{he_delving_2015} with Leaky ReLU (0.01 slope)~\cite{maas_rectier_2013}, and the second uses Xavier uniform initialization~\cite{glorot_understanding_2010}. Biases are initialized to zero. Using a batch size of four, we train each incremental task for two epochs. Images are resized to $224 \times 224$ and augmented via random horizontal flipping and cropping. We use the AdamW optimizer~\cite{loshchilov_decoupled_2018} with $10^{-4}$ weight decay. Per \cite{deng_zero-shot_2025}, the learning rate starts at $10^{-3}$ and reduces tenfold for the second epoch. The diversity loss weight is $\lambda_{\mathrm{div}} = 1$.

\section{Details on the Benchmarks}\label{sec:appendix:benchmark} 
D-RICO~\cite{neuwirth-trapp_rico_2025-2} aggregates 15 autonomous driving and surveillance datasets with diverse visual variations (sensor, lens, weather, viewpoint, domain). Focusing on three classes (\texttt{vehicle}, \texttt{bicycle}, \texttt{person}), it tests robustness to visual shifts and ambiguities. Tasks: daytime (daytm), thermal (therm), fisheye fix (fyfix), drone (drone), simulation (simul), fisheye car (fycar), RGB + thermal fusion (rgbth), video game (vgame), nighttime (night), fisheye indoor (fyind), gated (gated), photoreal. simulation (prsim), thermal fisheye indoor (tfyin), inclement (incle), event camera (evcam).

ODinW~\cite{li_elevater_2022} contains 13 datasets with partially overlapping vocabularies. Unlike D-RICO, it requires sequential learning of new concepts and classes, as well as handling semantic shifts. Tasks: Aerial Maritime Drone (Ae), Aquarium (Aq), Cottontail Rabbit (Co), EgoHands (Eg), Mushrooms (Mu), Packages (Pa), Pascal VOC (Pv), Pistols (Pi), Pothole (Po), Raccoon (Ra), Shellfish (Sh), Thermal Dogs and People (Th), Vehicles (Ve).

\section{Different Quantization and Success Probabilities}
\begin{figure}[t]
    \centering
    \begin{subfigure}[b]{0.42\linewidth}
        \centering
        \includegraphics[width=\linewidth, clip]{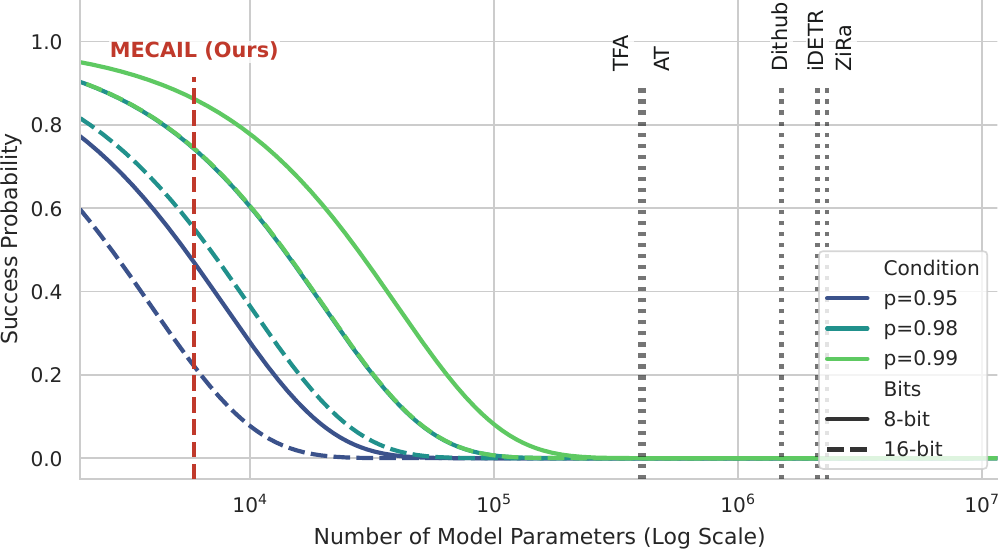}
        \caption{V2X success probability}
    \end{subfigure}
    \hfill 
    \begin{subfigure}[b]{0.42\linewidth}
        \centering
        \includegraphics[width=\linewidth, clip]{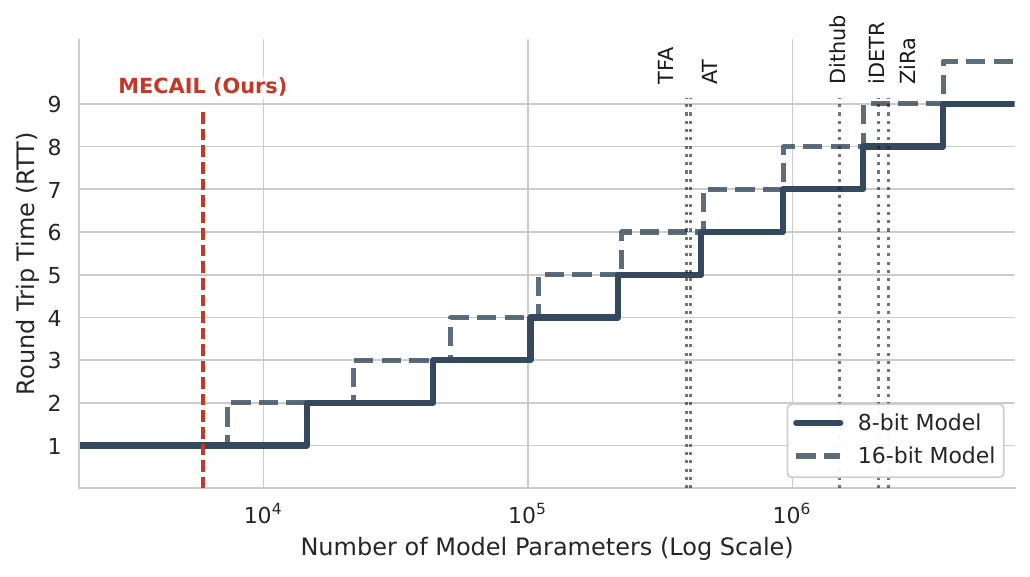}
        \caption{TCP RTT performance}
    \end{subfigure}

    \caption{\textbf{Performance comparison for different bit setups.}}
    \label{fig:combined_mecail_performance}
    \vspace{-19pt}
\end{figure}

Figure~\ref{fig:combined_mecail_performance} shows that 8-bit quantization fits more parameters into the initial TCP window. In V2X, this improves success probability or allows higher parameter counts.

\clearpage

\bibliographystyle{splncs04}
\bibliography{ref_addional, references} 

\end{document}